\documentclass[fleqn,10pt]{wlscirep}
\usepackage[utf8]{inputenc}
\usepackage[T1]{fontenc}

\usepackage{gensymb}
\usepackage{algorithm}
\usepackage{algpseudocode}
\usepackage{subfig}

\title{A New  Quaternion-Joint Cable-Driven Redundant  Manipulator Configuration and its Control Through FABRIK and Residual Reinforcement Learning}

\author{Tanapath Pornthisan, Thanapat Kemthong, Thanyapisit Kangsathien, Pasut Aranchaiya, Paulo Garcia, Viboon Sangveraphunsiri}
\affil{International School of Engineering, Chulalongkorn University, Bangkok, Thailand}

\affil{paulo.g@chula.ac.th}


\begin{abstract}
Robotic arms capable of traversing arbitrary spatial paths, especially in highly obstructed workspaces, are highly desired across several industries.
Quaternion-joints have recently empowered a specific class of robotic arms -cable-driven redundant manipulators- beyond its prior capabilities. Specifically, quaternion-joints reduce the number of required motors per degree of freedom, paving the way for more compact solutions.
An ongoing challenge is that the complexity of the kinematic model of quaternion joints challenges \textit{a priori} decisions on manipulator configurations and imposes higher computational demands on the control system and its non-linearities amplify all discrepancies between design and physical artifact arising from fabrication imprecision.
Here we show a that a 4-segment, 8-joint manipulator can achieve a broader workspace than extant configurations, at lower hardware cost, and that Residual Reinforcement Learning outperforms extant state-of-the-art methods -specifically, the FABRIK algorithm- on the control of such manipulator.
Our results show that this configuration is more workspace-effective than prior designs, and that Residual Reinforcement Learning outperforms FABRIK by three orders of magnitude on positional and orientational accuracy, effecting precise control of the novel 4-segment, 8-joint manipulator. Additionally, the control implementation is simpler: we describe the complete FABRIK process for control and corresponding learning implementation.
Our methodology is applicable to the design of new systems, providing designers with further tools for the development of this class of manipulators and corresponding control systems for novel configurations.

\end{abstract}
\begin{document}

\flushbottom
\maketitle
%
%
\thispagestyle{empty}

\section{Introduction}
Robot manipulators, both discrete and continuous \cite{cowan2013importance}, are employed across a wide range of applications to automate tasks traditionally performed by humans. Robotic arms have been used to automate satellite service and repair in space applications \cite{papadopoulos2021robotic}, assist with surgeries \cite{howe1999robotics}, factory automation \cite{ghodsian2023mobile}, etc. In situations with size and weight restrictions, traditional robotic arms are often constrained. Cable-driven redundant manipulators (CDRM) \cite{li2023force} offer better bending characteristics and adaptability while maintaining relatively simple kinematic derivation from its rigid links.  By using cables for remote motion transmission, CDRMs provide electromechanical separation between the drive components and the manipulator \cite{xu2018kinematics}, reducing the overall weight of the manipulator body. This design enables a higher length-to-radius ratio while maintaining flexibility and obstacle avoidance in constrained spaces. 
\par  CDRMs typically require 3N driving motors to supply 2N degrees of freedom (DOF). Recently, a CDRM design utilizing quaternion joints \cite{kim2018quaternion} achieved 2N motors for 2N DOF, offering wider bending angle and simpler structure compared to traditional methods. The ongoing challenge is that its control system is unsuitable for unstructured environments; efforts in the improvement of the control by introducing new methods such as forward and backward reaching inverse kinematics (FABRIK) \cite{aristidou2011fabrik} have been attempted,but require highly-precise fabrication to minimize kinematic discrepancies compared to the design (small manufacturing imprecisions can significantly degrade control accuracy \cite{ishiguro1992neural}), showing limited robustness in convergence to the desired configuration. 
\par Here we show an alternative mechanical design, characterized by fewer joints per segment but with wider bending angle per joint, and its control through  a conventional model-based controller and reinforcement learning with a residual term that compensates for modeling errors, cable hysteresis, and unmodeled dynamics, improving tracking accuracy and robustness.

\begin{figure}[!t]
\centering
\includegraphics[width=0.9\columnwidth]{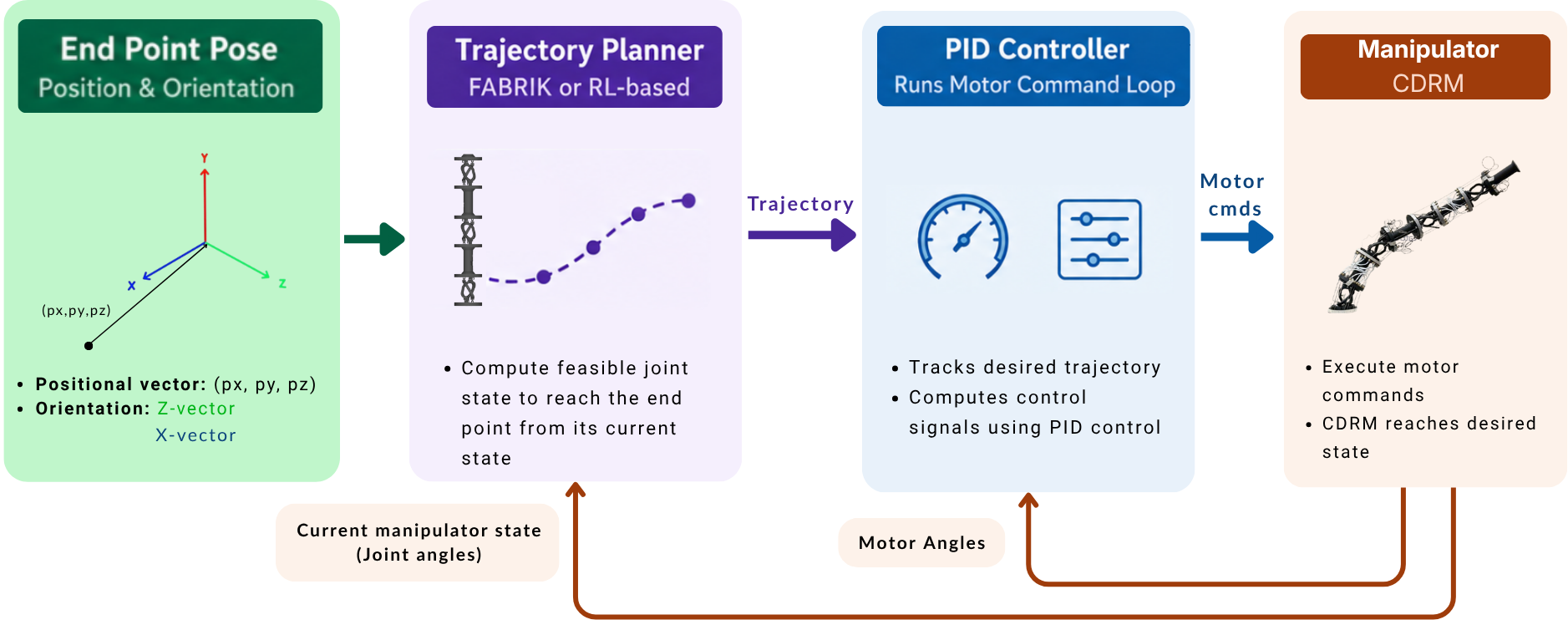}
\caption{Conceptual overview of the prototype system. Desired end-point pose can be estimated through either FABRIK or RRL, determining desired trajectory which is effected on the manipulator by the controller.}
\label{fig:res3}
\end{figure}

\par Specifically, this paper offers the following contributions:

\begin{itemize}
	\item We describe the mathematical model for a quaternion joint and the design and implementation of a prototype novel 4-segment/8-joint manipulator using said joints.
	\item We describe its control through inverse kinematics solving through the FABRIK algorithm.
	\item We describe the implementation of an equivalent Residual Reinforcement Learning (RRL) algorithm for the control of such manipulator.
	\item Experimental evaluation, on both simulated and physical implementations, is performed on workspace achievement, given maximum joint angles, and on the performance of the two controllers. 
\end{itemize}

\par The remainder of this paper is organized as follows: Section \ref{sec:model} describes the kinematic models of the utilized quaternion joints and implemented manipulator. Section \ref{sec:fabrik} describes its control through the forward and backward reaching inverse kinematics method, and Section \ref{sec:rrl} describes its control through the application of Residual Reinforcement Learning. Section \ref{sec:experiments} describes our experimental setup and methodology, as well as associated results. Section \ref{sec:related} positions the work within the broader state of the art, and Section \ref{sec:conclusions} concludes this work.

\section{Models}\label{sec:model}

\subsection{Modeling Quaternion Joints}

\begin{figure}[H]
    \centering
    \includegraphics[width=0.6\columnwidth]{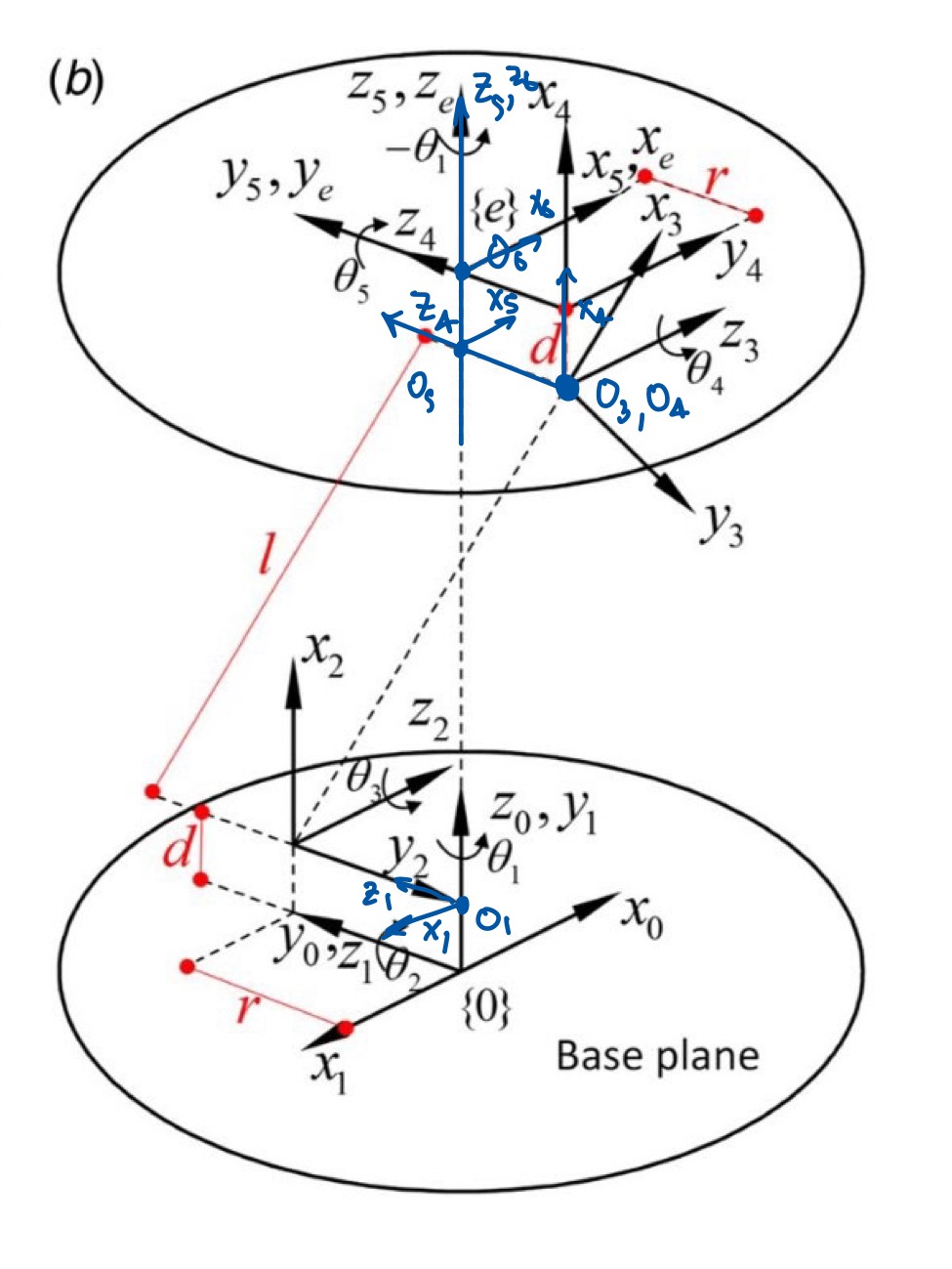}
    \caption{Denavit-Hartenberg model for the reference frames of a universal quaternion joint.}
    \label{fig:new_dh_model}
\end{figure}

The utilized quaternion joint, depicted in Fig. \ref{fig:new_dh_model} is modeled through the Denavit-Hartenberg (DH) parameters as an adaptation of the  model described in \cite{kim2018quaternion}. The origin of the first rotating joint $O_1$ is shifted upward a distance of $d$ so that the axis of rotation $Z_1$ intercepts with the axis of rotation of the second rotating joint $Z_2$. The origin of the third and forth rotating joint are in the same location, fully representing the universal joint's mechanism. To compensate for these changes, the frame of the 5th and 6th joint have been moved to satisfy the conditions of kinematic modeling, leading to the parameters depicted on  Table \ref{tab:dh_parameters}.

\begin{table}[htbp]
    \centering
    \caption{DH parameters of single middle limb of quaternion joint}
    \vspace{5pt}
    \label{tab:dh_parameters}
    \small
    \renewcommand{\arraystretch}{1.2} 
    \begin{tabular}{ccccc}
        \hline\hline
        $i$ & $a_i$/mm & $\alpha_i/^\circ$ & $d_i$/mm & $\theta_i/^\circ$ \\
        \hline
        1 & $0$ & $90$  & $d$ & $\theta_1 + 180$ \\
        2 & $0$ & $-90$ & $r$ & $\theta_2 + 90$ \\
        3 & $l$ & $0$   & $0$ & $\theta_3 + \arcsin(2r/l)$ \\
        4 & $0$ & $90$  & $0$ & $\theta_4 - \arcsin(2r/l)$ \\
        5 & $0$ & $90$  & $r$ & $\theta_5 + 90$ \\
        6 & $0$ & $0$   & $d$ & $-\theta_1$ \\
        \hline\hline
    \end{tabular}
\end{table}

A polynomial approximations for the bending direction $\varphi$ and bending angle $\theta$ results in: 

\begin{equation}
    \begin{cases}
        \varphi = \operatorname{atan2}(y_e, x_e) \\

        \theta = 2\operatorname{atan2}(\sqrt{x_e^2 + y_e^2}, z_e)
    \end{cases}
    \label{eq:quat_angles}
\end{equation}

\subsection{Modeling a 4-segment manipulator}

\begin{table*}[H]
    \centering
     \caption{Comparison of Manipulator Segment Configurations}
    \begin{tabular}{ | c | c | c | }
        \hline
        Segment & Huang et al\cite{huang2023sensing}  & Proposed Configuration \\
        \hline
        1 & 4 & 2 \\ 
        2 & 3 & 2 \\ 
        3 & 3 & 2 \\ 
        4 & 2 & 2 \\ 
        \hline
        Total & 12 & 8 \\
        \hline
    \end{tabular}
    \label{tab:proposedconfig}
\end{table*}

We modify the  segment configuration from the CDRM based on the configuration bu Huang et al \cite{huang2023sensing}.  Our proposed design reduces the number of joint in every segment to 2, resulting in 8 total joints instead of 12, as shown on Table 2. The forward kinematics are derived from the transformation matrices for an ideal spherical quaternion joint and a simple connecting link.  The transformation matrix for a single quaternion joint is defined as:

\begin{equation}\label{eq:T_joint}
    T_\text{joint}=\left[ \begin{matrix}
        1-2c^2_\phi s^2_{\psi/2} & -2s_{2\psi}s^2_{\psi/2} & c_\phi s_\psi & h_{\text{joint}}c_\phi s_{\psi/2} \\

        -2s_{2\psi}s^2_{\psi/2} & 1-2s^2_\phi s^2_{\psi/2} & s_\phi s_\psi & h_{\text{joint}}s_\phi s_{\psi/2} \\
        -c_\phi s_\psi & -s_\phi s_\psi & c_\psi & h_{\text{joint}}c_{\psi/2} \\
        0 & 0 & 0 & 1
    \end{matrix} \right]
\end{equation}

 where $\phi,\,\psi\,$ and $h_\text{joint}$ represent bending direction, bending angle, and joint height respectively.  The transformation matrix of the connecting link is:

\begin{equation}\label{eq:T_link}
    T_{\text{link}}= \left[\begin{matrix}
        1 & 0 & 0 & 0 \\
        0 & 1 & 0 & 0 \\
        0 & 0 & 1 & h_{\text{link}} \\
        0 & 0 & 0 & 1 
    \end{matrix}\right]
\end{equation}
 where $h_\text{link}$ being the height of the link.

\section{Control Mode: FABRIK}\label{sec:fabrik}

To serve as a baseline control method, the FABRIK algorithm was applied to the control of the proposed manipulator \cite{aristidou2011fabrik}. Given the desired end-point position of the manipulator ($P_d$), two vectors ($X_d$ and $Z_d$) which describe the desired end-point orientation, and the current state of the manipulator ($q_0$), the algorithm determines the closest configuration of the manipulator to the current state that satisfies the desired end-effector pose. This results in a 6-DOF problem where end point position and orientation are fully specified. To solve this problem, the algorithm is divided into three main components: virtual equivalent model conversion to convert the physical model to a vector model, inner iteration to solve five degrees of freedom, and outer iteration to solve the remaining single degree of freedom.

\par
\begin{figure}[H]
    \centering
    \includegraphics[width=0.7\columnwidth]{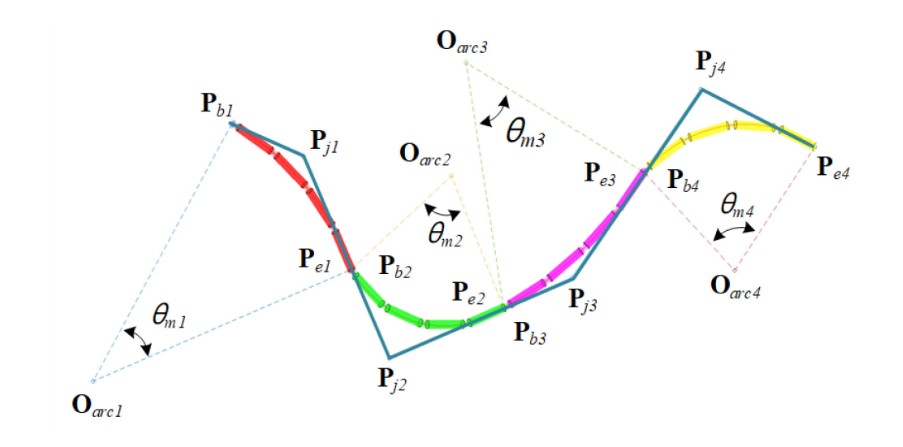}
    \caption{Equivalent model of the manipulator. Each segment is represented by three positional vectors and two directional vectors.}
    \label{fig:virtual_model}
\end{figure}

 The equivalent model of the manipulator is developed to support the FABRIK algorithm by describing its state through points in space as seen in Fig. \ref{fig:virtual_model}. This model represents each segment by three positional vectors and two directional vectors. The beginning of the segment is represented by positional vector $P_b$ and directional vector $Z_b$ (the blue line in Fig. \ref{fig:virtual_model}), while the end of the segment is represented by $P_e$ and $Z_e$. The remaining positional vector $P_j$ is the intersection of $Z_b$ and $Z_e$. 

Because of the decoupling design of the manipulator, each segment can be fully described by bending direction $\varphi$ and bending angle $\theta$ of a single quaternion joint. This design allows all quaternion joints inside a segment to have the same $\varphi$ and $\theta$, meaning the end point position and orientation of a segment can simply be obtained by repetitively multiplying the forward kinematics. After obtaining $Z_e$ from the forward kinematics, we can derive $P_j$ through the intersection between $Z_b$ and $Z_e$. Thus, we have fully derived the equivalent model. For multi-segment assemblies, kinematic continuity is preserved: the base position, $P_b$, and base tangent vector, $Z_b$, of the subsequent segment inherently coincide with the end position, $P_e$, and end tangent vector, $Z_e$, of the preceding segment, respectively.
The FABRIK algorithm also requires the derivation of length $P_{bk}P_{jk}$ and $P_{jk}P_{ek}$ as a function of $\theta_k$, where k is the k-th segment. The length of $P_{bk}P_{jk} = l_{k,1}$ can be expressed as: 

\begin{equation}
    l_{k,1}(\theta_k) = \frac{1}{2\cos\left(\frac{m_k\theta_k}{2}\right)} 
    \begin{cases} 

        K_{odd}, & \text{if } m_k \text{ is odd} \\ 
        K_{even}, & \text{if } m_k \text{ is even} 
    \end{cases}\\
    \label{eq:lk1}
\end{equation}

where

\begin{equation}
    K_{odd} =    h + 2h\sum_{i=1}^{\frac{m_k-1}{2}} \cos(i\theta_k) + 2l_0\sum_{i=1}^{\frac{m_k-1}{2}} \cos\left(\frac{2i-1}{2}\theta_k\right) 
\end{equation}

and

\begin{equation}
    K_{even} =  l_0 + 2l_0\sum_{i=1}^{\frac{m_k-2}{2}} \cos(i\theta_k) + 2h\sum_{i=1}^{\frac{m_k}{2}} \cos\left(\frac{2i-1}{2}\theta_k\right)
\end{equation}

 In addition, the length of $P_{jk}P_{ek} = l_{k,2}$ is expressed as: 

\begin{equation}
    l_{k,2}(\theta_k) = l_{k,1}(\theta_k) + l_0
    \label{eq:lk2}
\end{equation}

 where $l_0$ is the constant length of the catheter part.

\subsection{Inner Iteration}

\begin{figure}[H]
    \centering
    \includegraphics[width=0.8\columnwidth]{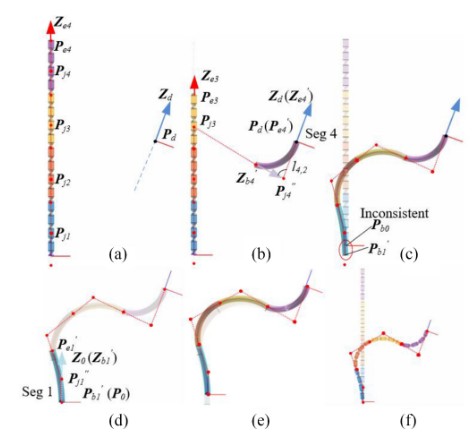}
    \caption{FABRIK innr iteration: (a) initial state; (b) Forward reach by segment number 4; (c) End of forward reach by segment number 4; (d) Backwards reach by segment number 1l (e) End of backwards reach; (f) State update.}
    \label{fig:fabrik_inner}
\end{figure}
 
The inner iteration of the FABRIK algorithm \ref{fig:fabrik_inner} is developed to solve 5 degrees of freedom from two inputs $P_d$ and $Z_d$ while considering the manipulator's current state, allowing the manipulator to reach the end point $P_d$ and align its final pointing direction to match $Z_d$, leaving the final end roll constraint to be resolved by the outer iteration. The inner iteration works by alternatively solving the forward stage and backward stage repetitively until positional error of the endpoints are acceptable. For the forward stage, we first set $P_e$ and $Z_e$ of the last segment of the equivalent model to $P_d$ and $Z_d$, respectively. We then find the new $P_j$ of the last segment through:

\begin{equation}
    \mathbf{P}'_{j4} = \mathbf{P}'_{e4} - l_{4,2}\mathbf{Z}'_{e4} = \mathbf{P}_d - l_{4,2}\mathbf{Z}_d
\end{equation}

\noindent Then, we calculate the new $Z'_{b4}$ which points from $P_{j3}$ to $P'_{j4}$: 

\begin{equation}
    \mathbf{Z}'_{b4} = \frac{\mathbf{P}'_{j4} - \mathbf{P}_{j3}}{\left\| \mathbf{P}'_{j4} - \mathbf{P}_{j3} \right\|}
\end{equation}

\noindent As $Z'_{e4}$ and $Z'_{b4}$ changes, $\theta_4$ has also shifted, compensating the manipulator's end to be $P_d$ and $Z_d$. Therefore, $\theta'_4$ can be expressed as:

\begin{equation}
    \theta'_4 = \frac{\arccos(\mathbf{Z}'_{e4}, \mathbf{Z}'_{b4})}{4}
\end{equation}

\noindent Because $\theta$ of the last segment changes, $l_{4,1}$ and $l_{4,2}$ has also been changed. Thus, $l'_{4,1}$ and $l'_{4,2}$ are recalculated. Next, $P'_{j4}$ and $P'_{b4}$ are also adjusted using the derived length:

\begin{equation}
    \mathbf{P}'_{j4} = \mathbf{P}'_{e4} - l'_{4,2}\mathbf{Z}'_{e4}
    \label{eq:update_pj4}
\end{equation}

\begin{equation}
    \mathbf{P}'_{b4} = \mathbf{P}'_{j4} - l'_{4,1}\mathbf{Z}'_{b4}
    \label{eq:update_pb4}
\end{equation}

\noindent Forward reaching of the last segment can be seen in Fig. \ref{fig:fabrik_inner}. For all segments, the procedure repeats iteratively from the last segment to the base, where $P'_{bk} = P'_{e(k-1)}$ and $Z'_{bk} = Z'_{e(k-1)}$. After reaching the base, we will see that $P'_{b1}$ is not aligned with the actual base location at [0,0,0]. Therefore, while the forward stage tries to adjust the end point of the manipulator and its direction to be $P_d$ and $Z_d$, the backward stage tries to map the position and direction of the base back to its original position ($P_{b1}$ at [0,0,0] and $Z_{b1}$ of [0,0,1]). To achieve this, backward reaching follows the similar procedure as the forward stage, but iteratively repeats from the base to the end of the manipulator.

\subsection{Outer Iteration}
Following the 5-DOF inner iteration, the outer iteration resolves the final remaining degree of freedom associated with the manipulator's roll angle. By introducing the desired end-effector X-axis, $\mathbf{X}_d$, as an additional constraint, the algorithm computes the rotation required around the Z-axis to fully align the manipulator with the target orientation. It works by first computing the roll error: 

\begin{equation}
    \psi = acos(\mathbf{X}_{\text{end}}, \mathbf{X}_d)
    \label{eq:roll_error}
\end{equation}

We then rotate the entire manipulator around the axis $\mathbf{Z}_d$ by an angle of $\lambda\psi$ (where $\lambda$ acts as a scaling or relaxation factor), aiming to align the X vector of the end-effector with $\mathbf{X}_d$. This behavior is applied using the matrix formulation of Rodrigues' rotation formula. First, we compute the rotation matrix $\mathbf{R}$ using the skew-symmetric matrix $\mathbf{A}^*$ of the normalized rotation axis $\mathbf{Z}_d$:

\begin{equation}
    \mathbf{R} = \mathbf{I} + \sin(\lambda\psi) \mathbf{A}^* + (1 - \cos(\lambda\psi)) (\mathbf{A}^*)^2
    \label{eq:rodrigues_matrix}
\end{equation}

\noindent By applying this rotation matrix, we can find the updated positions of the points in the virtual equivalent arms, $\mathbf{P}'_x$, relative to the rotation center $\mathbf{P}_d$. The rotation is applied to each node through:

\begin{equation}
    \mathbf{P}'_x = \mathbf{P}_d + \mathbf{R} (\mathbf{P}_x - \mathbf{P}_d)
    \label{eq:rodrigues_roll}
\end{equation}

\noindent Following the roll rotation, a positional discrepancy emerges because the manipulator's base significantly shifts away from its original physical location. To resolve this unanchored state, the rotated positions of the virtual equivalent arms must re-enter the inner iteration. This stage re-anchors the base to the correct origin while simultaneously converging the end-effector's X-axis toward the desired directional vector, $\mathbf{X}_d$. This two-layer iterative process repeats until the end-effector pose converges within the predefined error tolerance or the maximum number of allowable iterations is exceeded.

\section{Control Model: Residual Reinforcement Learning}\label{sec:rrl}
To generate a suitable base controller that accurately reflects the physical constraints of the new joint design, we implemented an FK-based data generation pipeline. New training data is acquired by systematically sampling the 8 cable lengths, simulating the actions in a physics-accurate simulation environment (MuJoCo \cite{todorov2012mujoco}) until the CDRM mechanism settles into a steady state (defined by mean joint velocity falling below a strict threshold), and then saving the resulting 9-dimensional (9D) endpoint pose data. This dataset captures the true nonlinear mapping from cable space to task space.

Instead of standard Behavior-Cloning, the baseline posture is provided by a deep Forward Model (FM) combined with an inference-time Inverse Kinematics (IK) solver. 

The Forward Model is parameterized by a neural network with the following architecture:
\begin{itemize}
    \item \textbf{Input Layer \& Feature Engineering:} The network takes the 8 absolute cable lengths (acting as positional deltas relative to home). The input is scaled and augmented with a 6-frequency Fourier feature encoding (concatenating sine and cosine projections) to better capture high-frequency task-space details.
    \item \textbf{Hidden Architecture:} The features pass through an initial dense layer with 512 hidden units, followed by 6 sequential Residual Blocks. Each block consists of two 512-dimensional dense layers. This is then passed through a Gaussian Error Linear Unit (GELU) activation layer.
    \item \textbf{Output Layer:} A final linear dense layer outputs the predicted 9D end-effector pose.
\end{itemize}

During execution, a JAX-optimized IK solver uses this Forward Model to find the target cable lengths. For a given 9D target pose, the solver employs gradient descent (via the Adam optimizer) over 200 iterations with a learning rate of $0.05$ and 8 random restarts. It optimizes a custom loss function that minimizes both positional and rotational prediction errors, alongside a small control regularization penalty.

While the FK Base Model provides a strong nominal posture, the physical manipulator exhibits predictable, systematic kinematic errors when operating at extended bending angles. To mitigate these nonlinearities, an RL residual policy adds small corrective adjustments directly to the base controller's output. The RRL controller learns these systematic kinematic deviations and adapts the joint commands to maintain high precision for the end-effector.

The RL residual policy is trained using an asymmetric actor-critic architecture within the Brax framework, which fundamentally distinguishes between the limited information available during physical deployment and the privileged information available during simulation.

\begin{itemize}
    \item \textbf{Actor Network (Policy Observation)}: The actor strictly limits its observation space to what the real robot can reliably produce. This includes the processed sensor history (estimated bend angle and direction for each segment, derived via polynomial fitting and forward kinematics), the bounded action history over the last 4 frames, the 9D target pose, and the baseline cable output from the IK solver.
    
    \item \textbf{Critic Network (Value Observation)}: The critic operates exclusively during training and receives additional, true state parameters appended directly to the actor's observations: the exact tendon lengths, tendon velocities, true joint positions, true 9D end-effector pose, and explicit current and initial positional and angular error metrics.
\end{itemize}

To ensure stability, the high-level control loop operates at a frequency of 50 Hz. Meanwhile, the MuJoCo physics engine simulates the manipulator dynamics at a much finer resolution. 

To bridge the simulation-to-reality gap, the simulation applies extensive Domain Randomization (DR) during the RRL training. We specifically randomize cable viscous damping (creep) and Coulomb friction, joint rotor inertia, actuator position gains, and structural mounting orientation (via a gravity vector perturbation). Mass and inertia terms, however, are deliberately preserved to maintain internal physics consistency within the simulated solver.

\section{Experiments and Result}\label{sec:experiments}

\subsection{Experimental Setup}

\begin{figure*}[!t]
\centering
\subfloat[]{
	\includegraphics[width=0.23\textwidth]{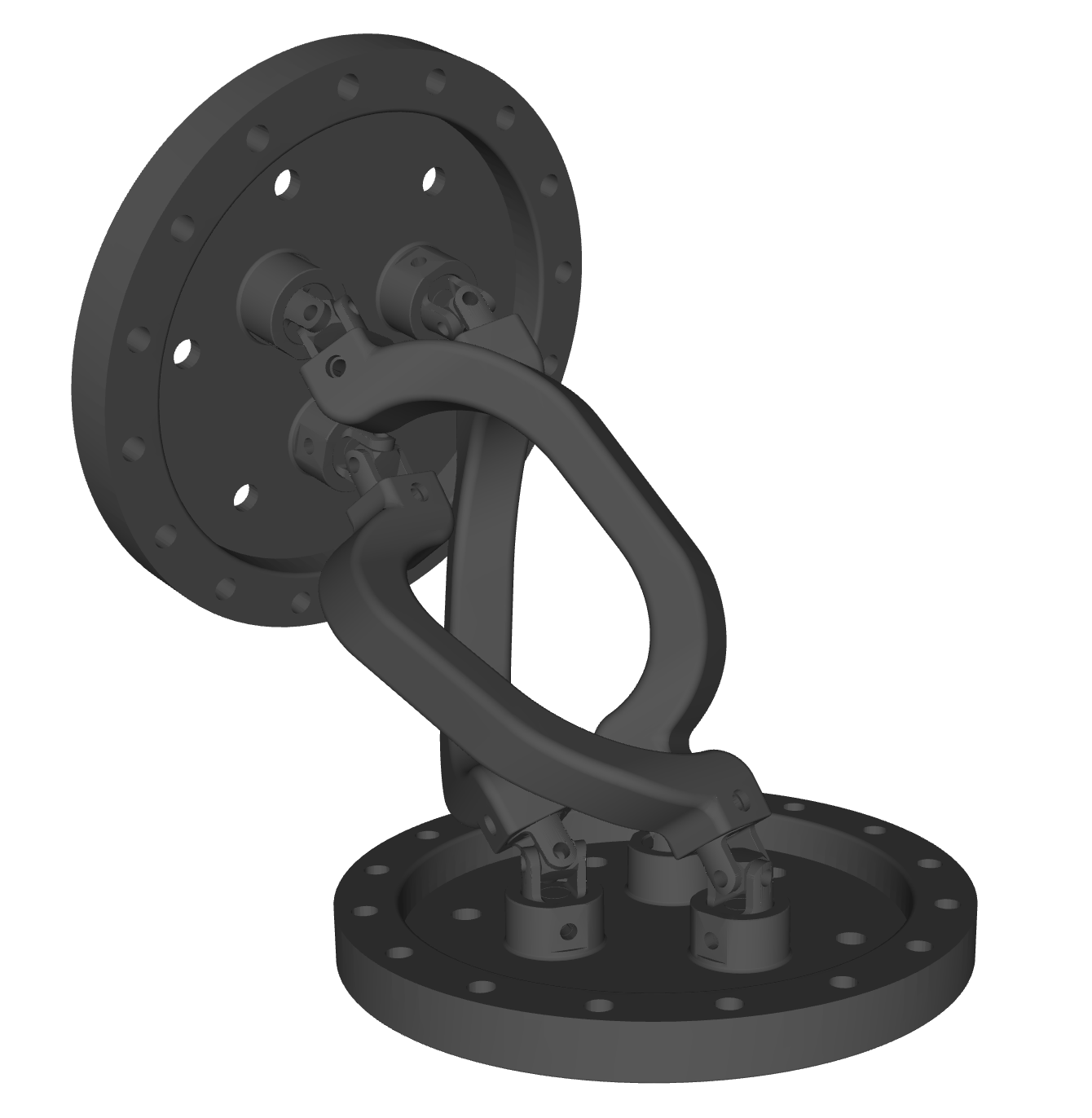}}
\subfloat[]{
	\includegraphics[width=0.18\textwidth]{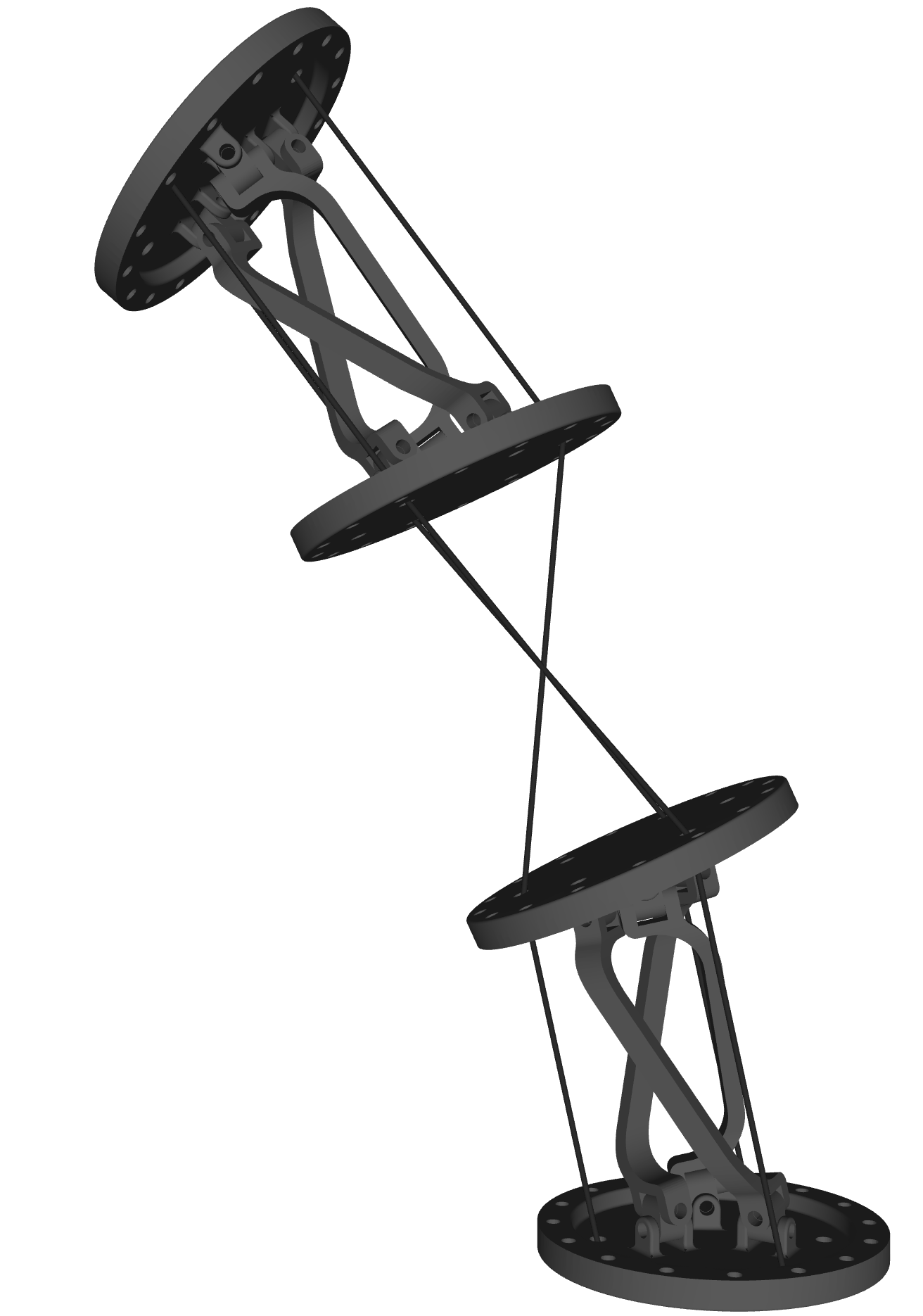}}
\subfloat[]{
	\includegraphics[width=0.13\textwidth]{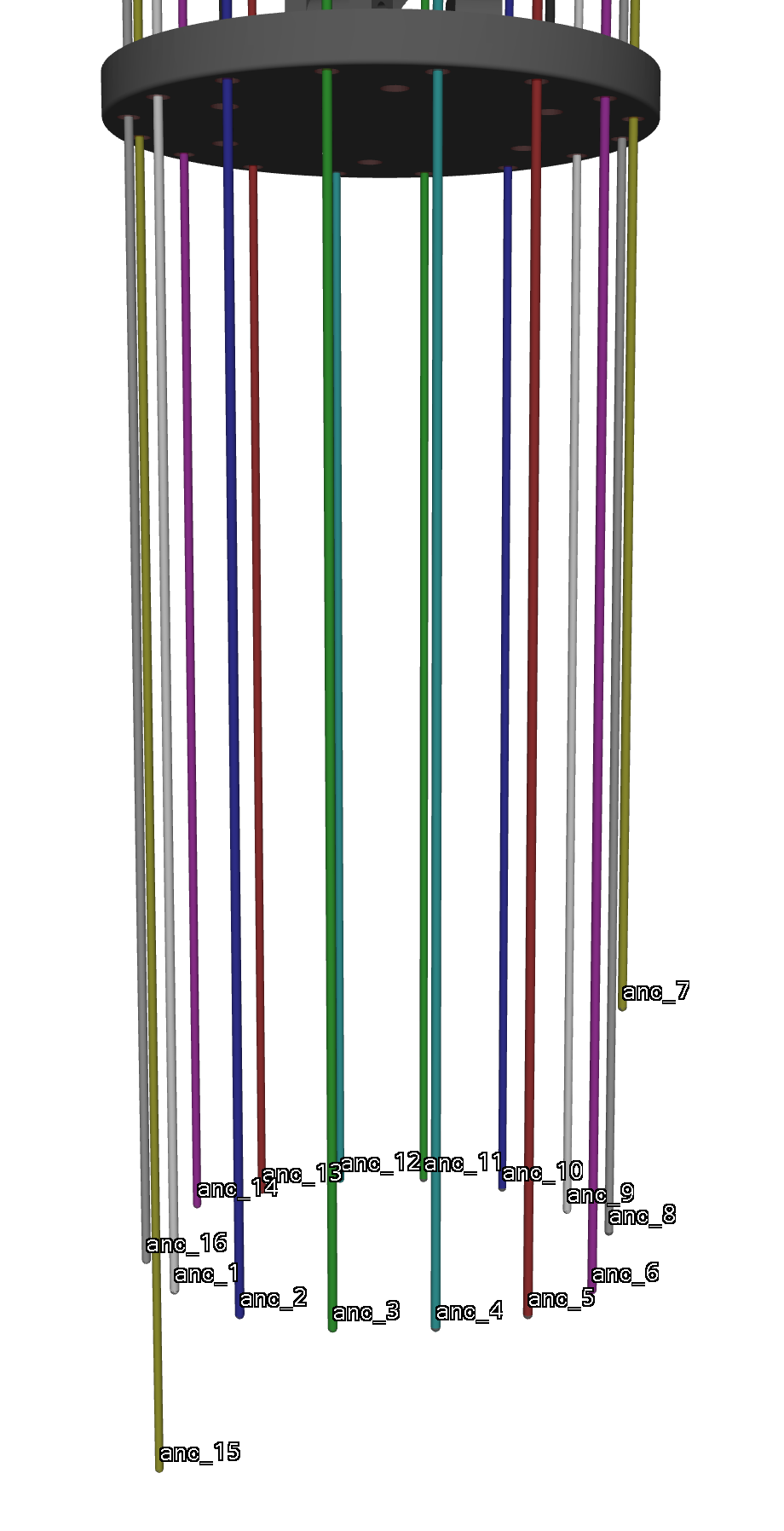}}
\subfloat[]{
	\includegraphics[width=0.25\textwidth]{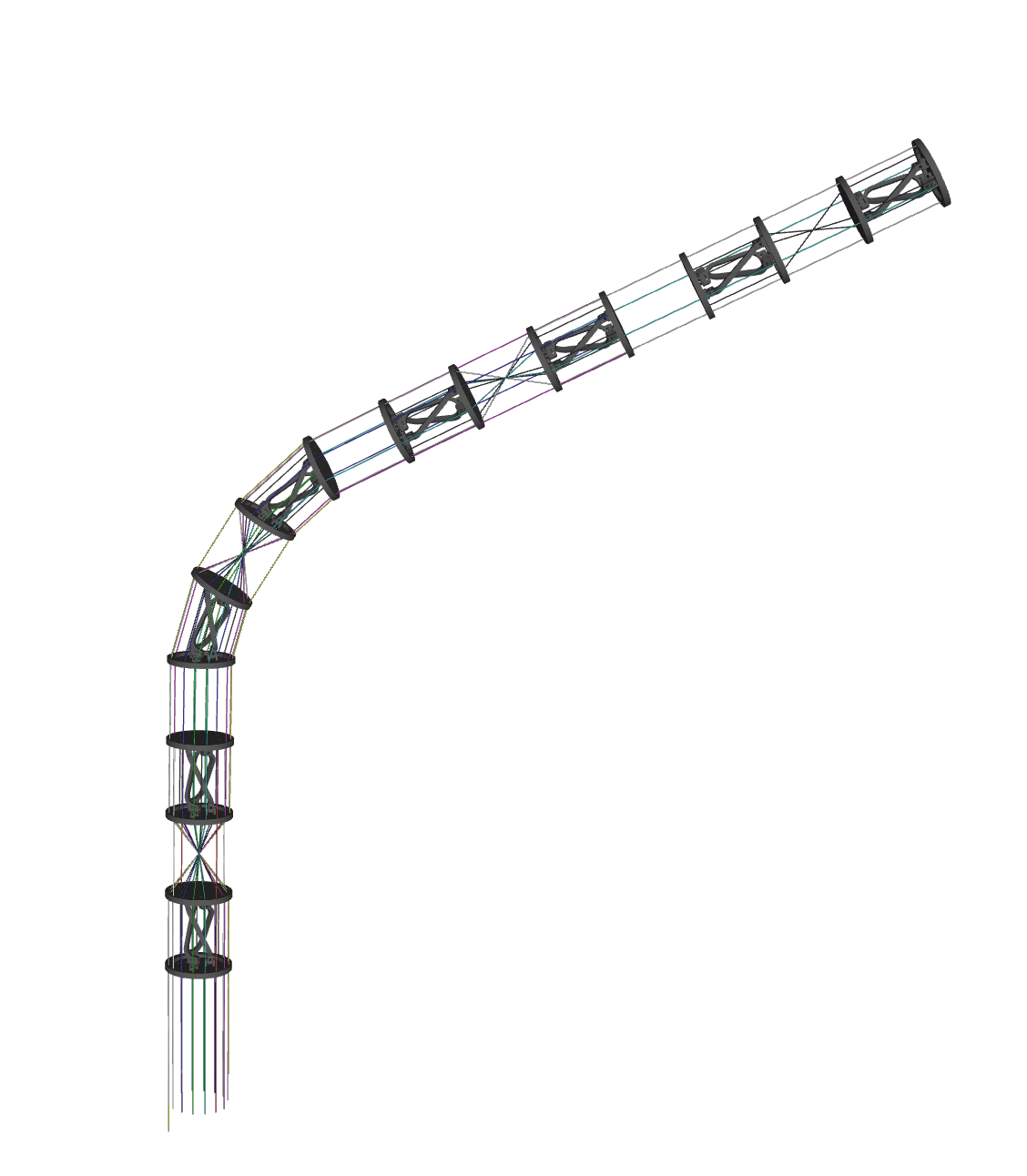}}
\caption{Digital (simulated) implementation of the manipulator. (a) 3D-model of the quaternion joint; (b) Joint linkage; (c) Cable connection through a segment; (d) Full 4-segments/8-joints manipulator.}
\label{fig:sim}
\end{figure*}

\begin{figure*}[!t]
\centering
\subfloat[]{
	\includegraphics[width=0.18\textwidth]{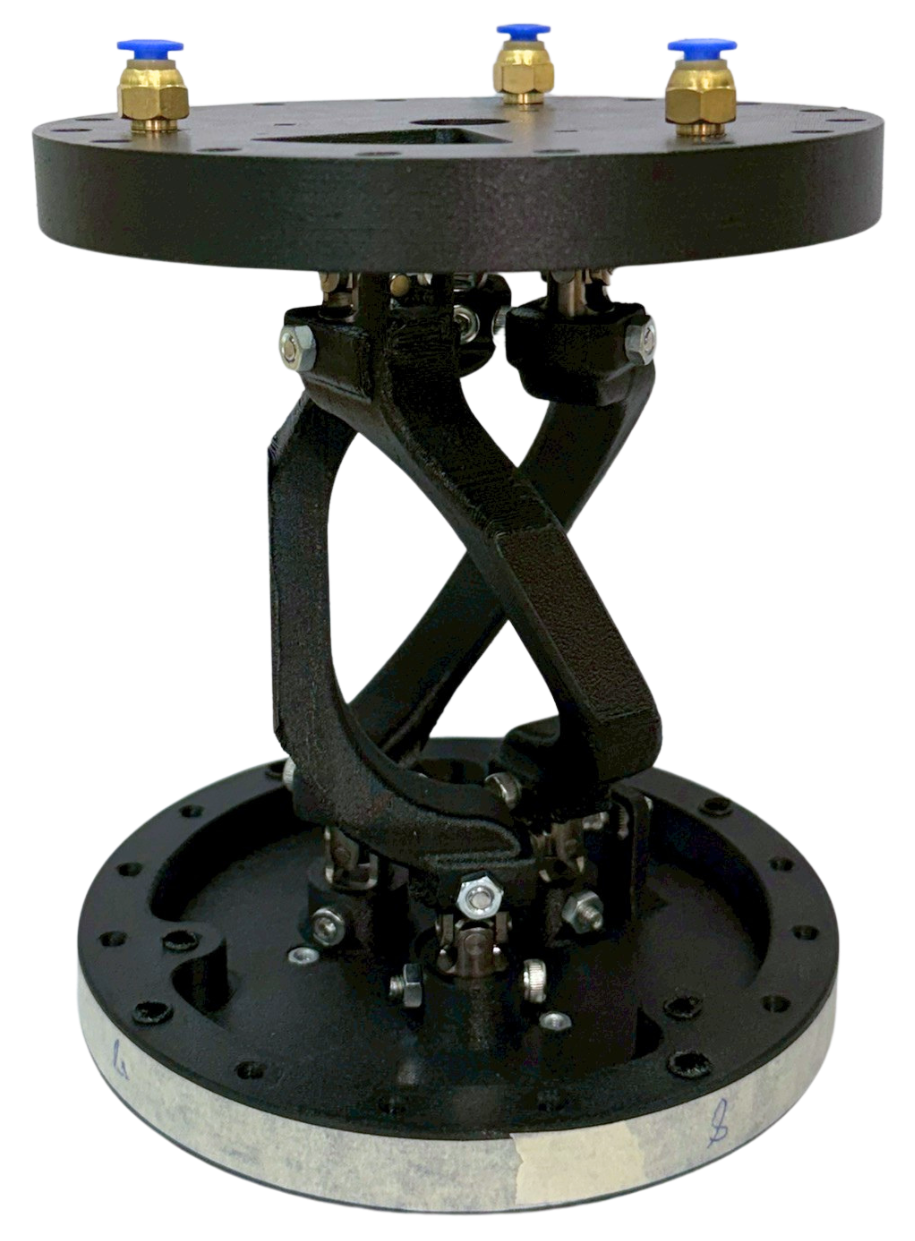}}
\subfloat[]{
	\includegraphics[width=0.18\textwidth]{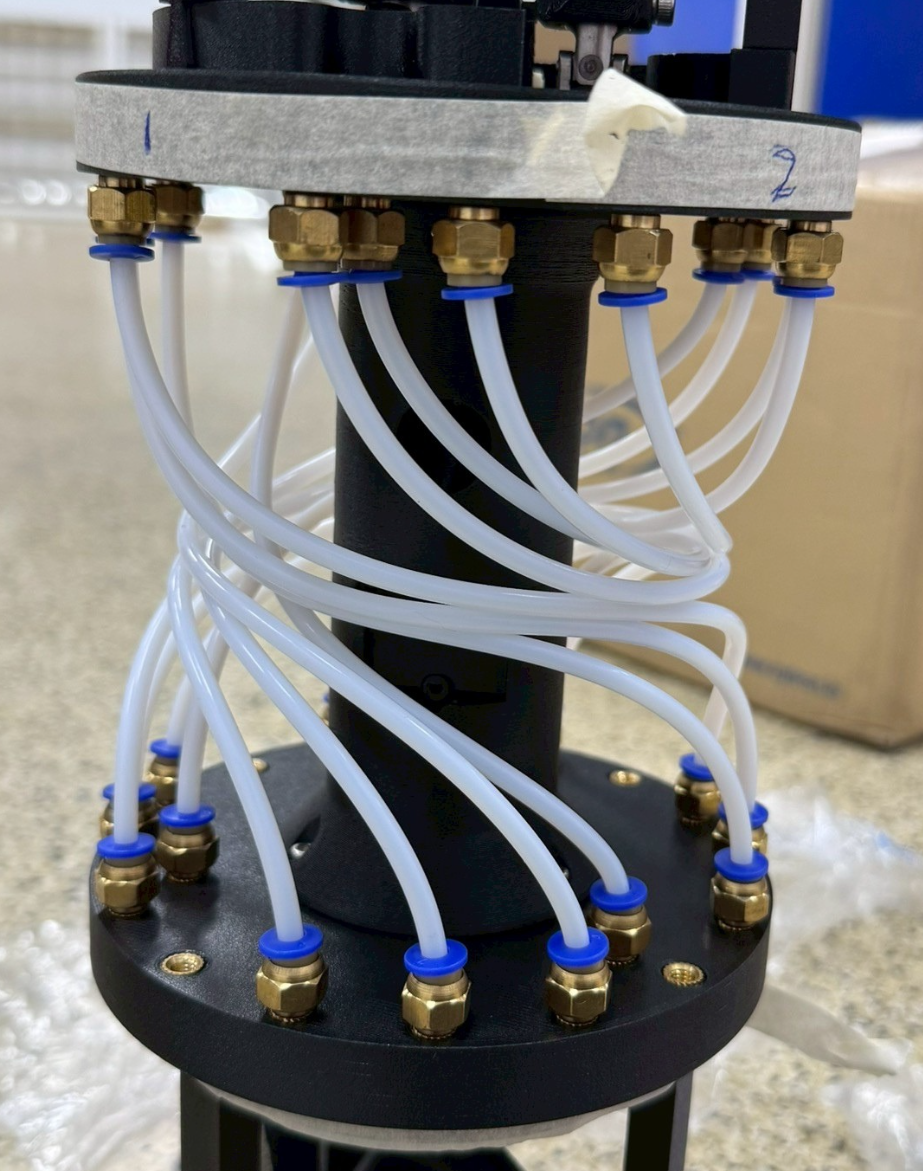}}
\subfloat[]{
	\includegraphics[width=0.31\textwidth]{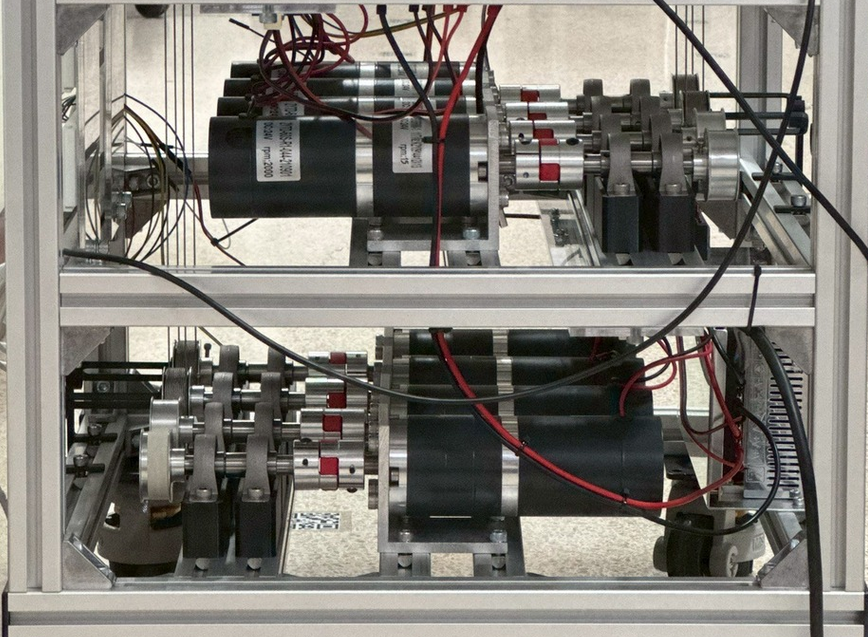}}
\subfloat[]{
	\includegraphics[width=0.17\textwidth]{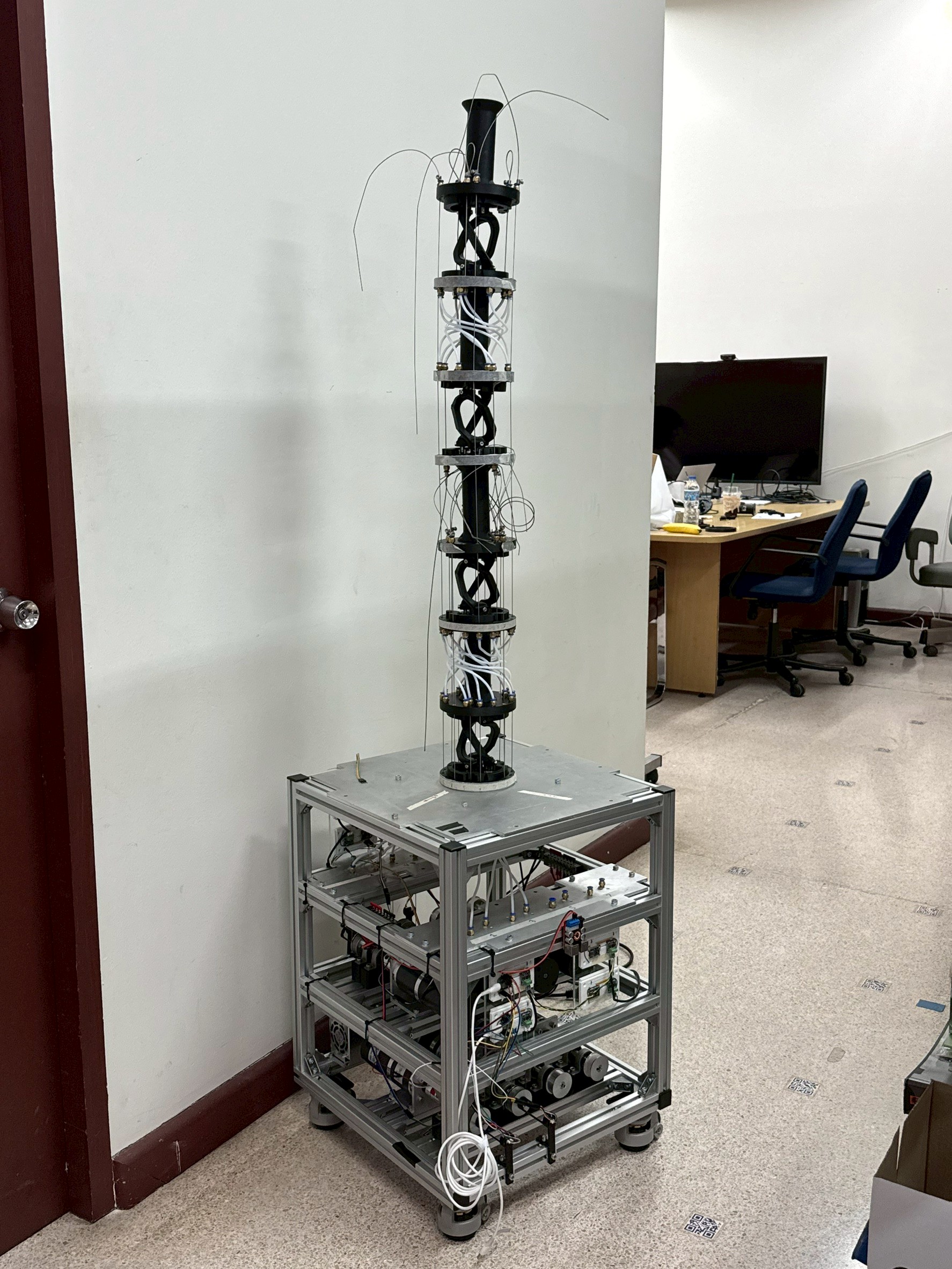}}
\caption{Physical implementation of the manipulator. (a) 3D-printed quaternion joint; (b) Cable-connections across a joint; (c) Drive-box; (d) Full-protptype implementation (2-segments).}
\label{fig:phys}
\end{figure*}

We have developed a virtual implementation which allowed us to explore the design space in simulation (Fig. \ref{fig:sim}), and a physical implementation (Fig. \ref{fig:phys}). Simulation was effected on Mujoco \cite{todorov2012mujoco}, primarily due to its native support for spatial tendons. In the physical implementation, each degree of freedom (corresponding to one cable) is effected through a shaft, two bearing blocks, a motor mount, pulley, and coupling. We use a ZGX60RMM  with 4.9Nm torque and 15rpm, with a 40mm-diameter pulley, KP001 bearings, and D30L40 couplings. Quaternion joints were 3D-printed in carbon-fiber reinforced PETG using Fused Deposition Modeling. The final prototype consists of 4 segments, mounted on a controller base that houses mechanical drive, sensing, and actuation electronics.

\subsection{Fabrik validation}

We validate our implementation of the FABRIK algorithm for determining the end-point position and orientation of the multi-segment manipulator. To establish a direct comparison, we adopted the testing methodology outlined in \cite{huang2023sensing}. We generated a dataset of 1,000 random ground-truth poses by sampling joint angles within the ranges of 0Â° to 360Â° for $\varphi$ and -30Â° to 30Â° for $\theta$. Each data point comprises four distinct sets of these joint variables (one for each segment), which were then processed through our forward kinematics model to yield the target end-effector poses. The FABRIK algorithm was subsequently evaluated by initializing it from a neutral home configuration and tasking it to converge on each of the 1,000 target poses. To clarify, the algorithm is considered successful if the resulting end effector is within 0.01mm positional error and 0.01Â° roll error measured by the X vector. We compared our approach against the baseline  based on overall success rate and average computation time. To contextualize the timing results, all computational benchmarks were executed in MATLAB on a 64-bit Windows machine equipped with an AMD Ryzen 7 5800H processor (3.20 GHz) and 16 GB of RAM.

\begin{table}[htbp]
    \centering
    \caption{FABRIK comparison}
    \vspace{5pt}
    \label{tab:fab_comparison}
    \renewcommand{\arraystretch}{1.0} 
    \begin{tabular}{ccc}
        \hline
        \textbf{Method} & \textbf{Success rate(\%)} & \textbf{Average computation time (s)} \\
        \hline
        \cite{huang2023sensing} & 77.6 & 0.3705 \\
        \\
        Ours & 70.4 & 0.0221\\
        \\
        \hline
    \end{tabular}
\end{table}

To convert the results of the FABRIK algorithm to physical cable length actuation, the core geometric model is derived below based on the system of equations for each segment $i$:

\begin{equation}
    \begin{cases} 
        m_i D \cos(\beta_1^i - \varphi_i) \sin(\theta_i/2) = l_1^{S_i} \\ 
        m_i D \cos(\beta_2^i - \varphi_i) \sin(\theta_i/2) = l_2^{S_i} \\ 
        \beta_2^i = \beta_1^i + (\pi/2) 
    \end{cases}
    \label{eq:kinematics}
\end{equation}

\noindent This formulation scales the segment's maximum structural displacement limit ($m_i D$) by the desired bending magnitude ($\sin(\theta_i/2)$) and projects it onto the specific physical routing angles ($\beta_1^i, \beta_2^i$) of each opposing cable pair, relative to the bending plane angle ($\varphi_i$). To actuate the physical hardware, the software explicitly maps these calculations to specific orthogonal cable pairs for each segment based on their hardware offsets: 
\begin{itemize}
    \item \textbf{Segment 1:} Cables 1 at $0^\circ$ and 5 at $90^\circ$
    \item \textbf{Segment 2:} Cables 3 at $45^\circ$ and 7 at $135^\circ$
    \item \textbf{Segment 3:} Cables 2 at $22.5^\circ$ and 6 at $112.5^\circ$
    \item \textbf{Segment 4:} Cables 4 at $67.5^\circ$ and 8 at $157.5^\circ$
\end{itemize}

\begin{figure*}[!t]
\centering
\subfloat[]{
	\includegraphics[width=0.24\textwidth]{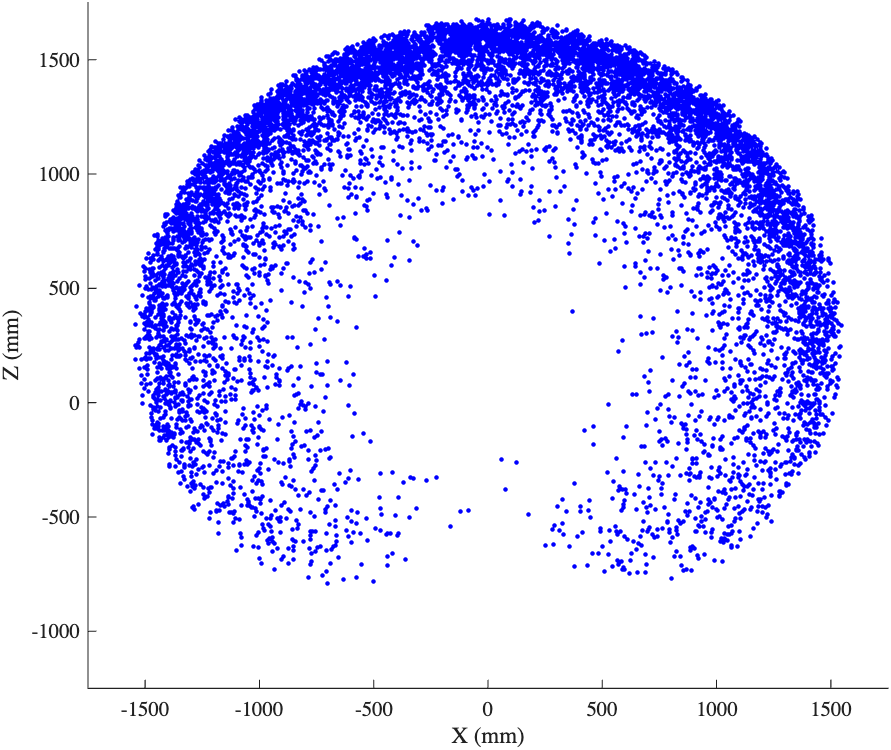}}
\subfloat[]{
	\includegraphics[width=0.24\textwidth]{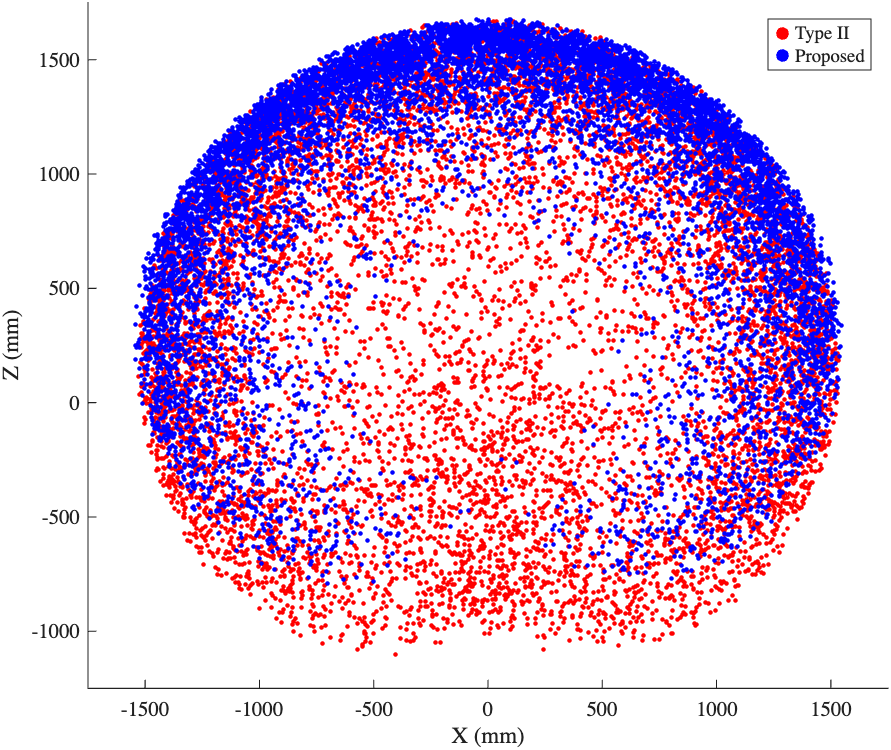}}
\subfloat[]{
	\includegraphics[width=0.24\textwidth]{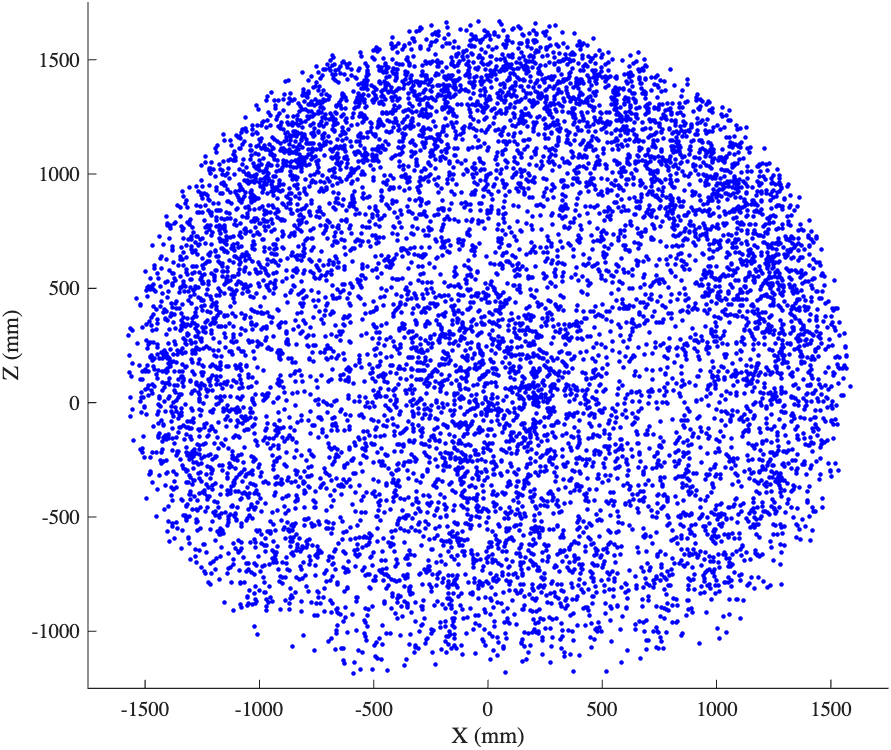}}
\subfloat[]{
	\includegraphics[width=0.24\textwidth]{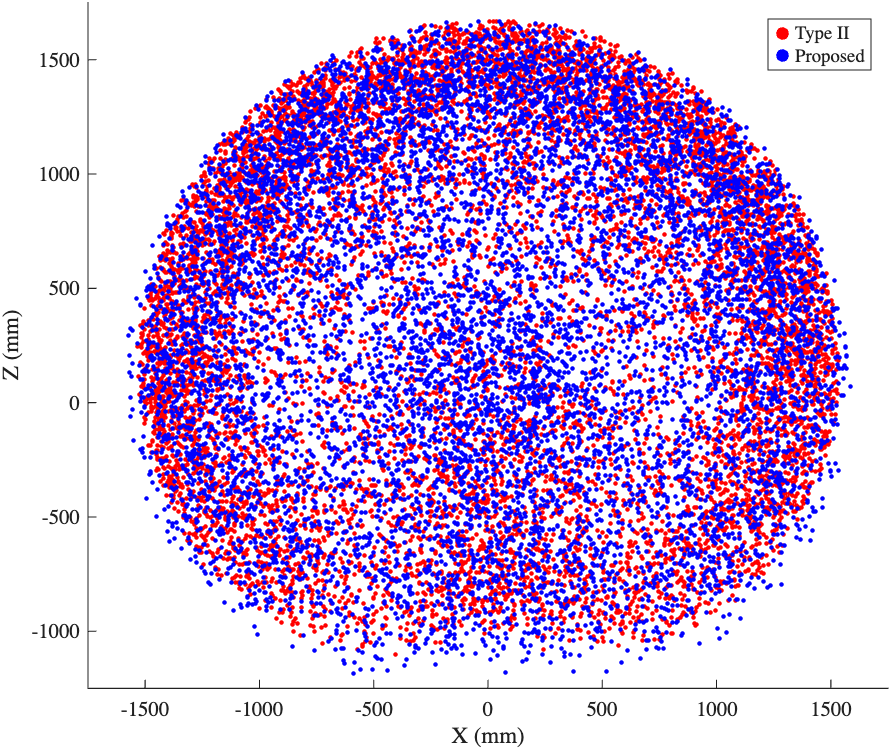}}
\caption{MCMC-simulation of achieveable workspace. (a) Estimated workspace (40$\degree$ angle); (b) 40$\degree$ angle workspace (blue) overlayed on workspace by \cite{} (red);  (c) Estimated workspace (70$\degree$ angle); (d) 70$\degree$ angle workspace (blue) overlayed on workspace by \cite{} (red); }
\label{fig:mcmc}
\end{figure*}

\begin{figure*}[!t]
\centering
\includegraphics[width=0.9\textwidth]{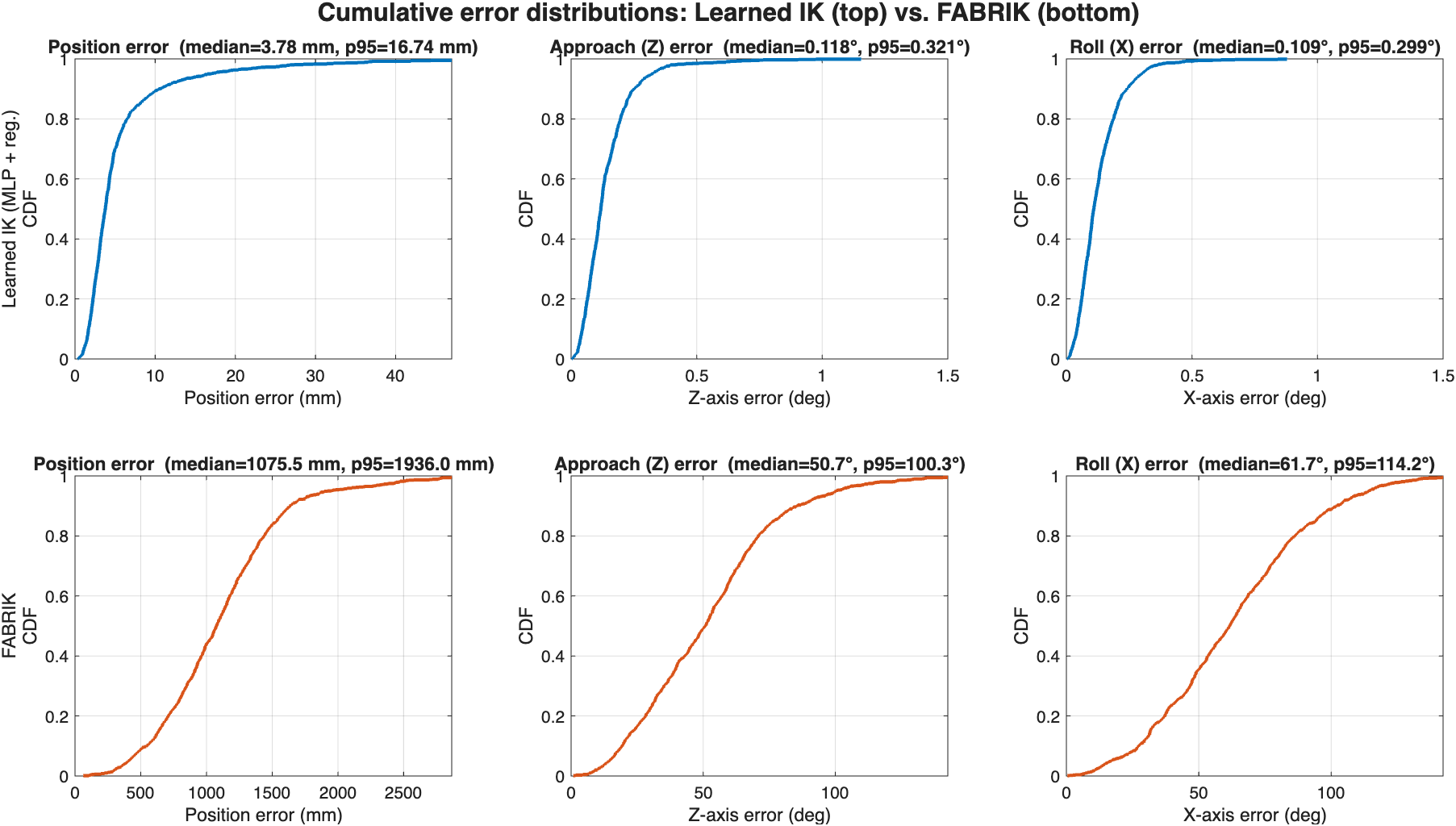}
\caption{Cumulative error distributions for FABRIK and RRL (Learned IK).}
\label{fig:rrl}
\end{figure*}

\subsection{Workspace analysis}
We first simulated the proposed configuration using the original joint angle range of $[-40\degree,\,40\degree]$. The resulting workspace is shown in Fig.\ref{fig:mcmc}. However, when compared with Huang et al's \cite{huang2023sensing} configuration, our proposed configuration clearly has a smaller reachable area than the original. We ran a second simulation increasing the maximum joint bending angle to $70\degree$. \\

\begin{figure}[!t]
\centering
\subfloat[]{
	\includegraphics[width=0.65\columnwidth]{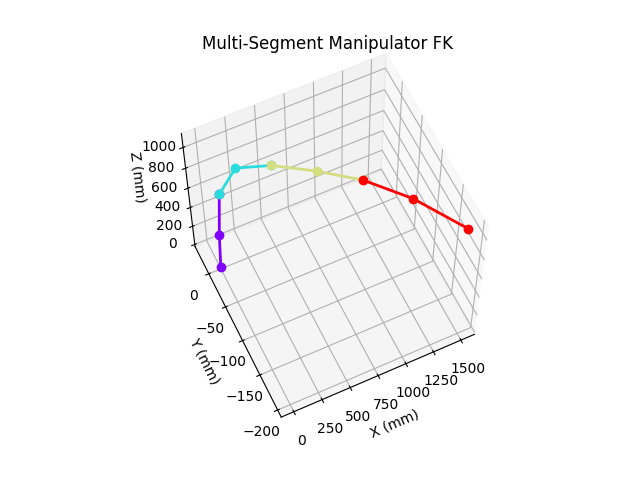}}
\subfloat[]{
	\includegraphics[width=0.34\columnwidth]{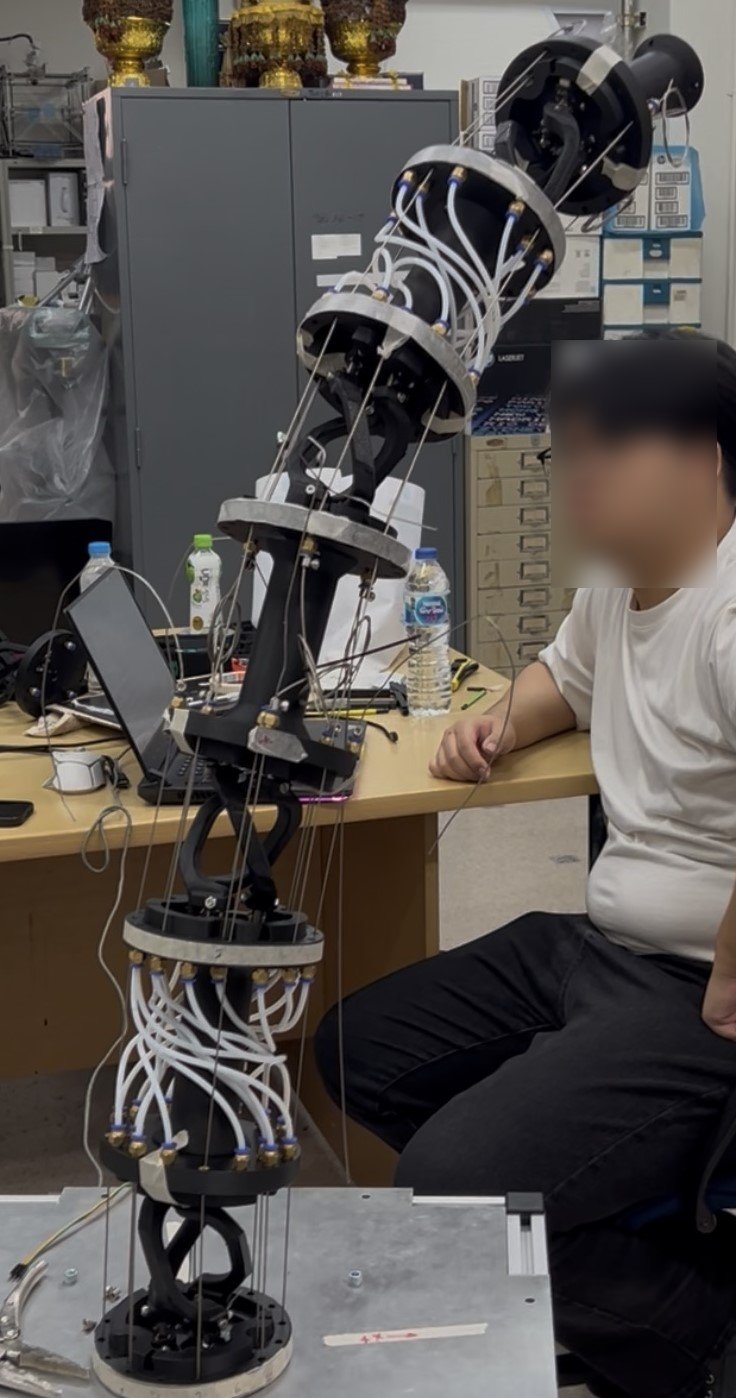}}
\caption{Multi-segment forward kinematics test. (a) numerical simulation; (b) physical validation.}
\label{fig:demo}
\end{figure}

The results of this modification are highly successful.  Fig. \ref{fig:mcmc} provides a direct comparison, demonstrating that the proposed configuration with a $70\degree$ maximum angle in blue achieves a comparable workspace. This result validates that our simplified 8-joint design can achieve the target workspace by increasing the operational range of each joint.
The modification demonstrates that the proposed configuration maintains the desired workspace while offering several practical advantages. By reducing the number of joints, the mechanical design becomes simpler, faster to fabricate, and more cost-efficient, making the overall system more feasible within the project timeline. At the same time, increasing the joint angle range effectively compensates for the reduced joint count, allowing the manipulator to preserve comparable workspace coverage and motion flexibility. A visual representation of control is depicted n Fig. \ref{fig:demo}.

\subsection{Residual Reinforcement Learning}
\subsubsection{MJX Joint Velocity Jitter}
Following the development of a suitable base controller, our focus shifted to training the residual policy. However, during the initial training phase, a significant numerical instability was identified within the Reinforcement Learning (RL) environment.

To optimize training efficiency, MuJoCo MJX was utilized to execute parallel simulations directly on the GPU. By co-locating both the simulation engine and the RL model on the GPU, unnecessary memory transfers typically associated with CPU-based simulations are theoretically eliminated. Nevertheless, the MJX implementation exhibits subtle computational differences compared to the traditional CPU-based MuJoCo engine. Specifically, we observed that enforcing tight tendon length boundaries induces high-frequency jitter in the joint velocities. This poses a major issue, as these velocities are essential for generating the target bank and serve as critical state inputs to the critic network. Fig. \ref{fig:rrl}  illustrate the disparity in the steady-state joint velocities under identical control commands for CPU and GPU simulations, respectively.

Extensive parameter tuning was conducted to address this instability, including relaxing the  parameters and adjusting the solver iterations and tolerances. The only modification that successfully eliminated the velocity jitter was expanding the tendon length range. However, this adjustment compromised the CDRM ability to accurately hold its position, as the system must be modeled with rigid cables rather than highly flexible tendons. 

Initial mitigation efforts involved applying a moving average filter to smooth the joint velocities; however, this approach failed to yield satisfactory training results. Ultimately, to bypass the MJX instability without sacrificing the rigidity of the cables, the target bank was generated using the standard CPU implementation, and the joint velocities were entirely omitted from the critic network's observation space.

\begin{figure*}[!t]
\centering
\subfloat[]{
	\includegraphics[width=0.9\textwidth]{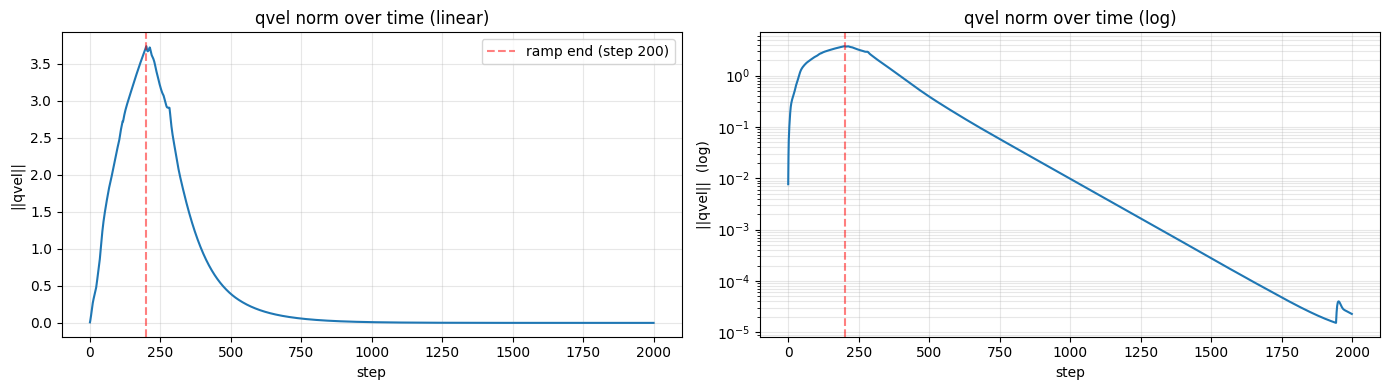}}\\
\subfloat[]{
	\includegraphics[width=0.9\textwidth]{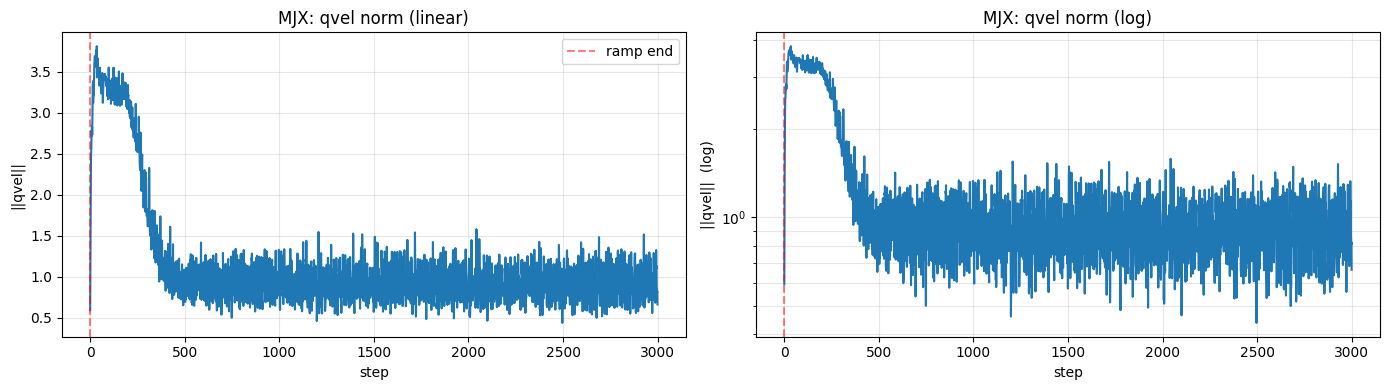}}
\caption{Residual Reinforcement Learning convergence (qvel norm); (a) CPU; (b) GPU.}
\label{fig:res2}
\end{figure*}

\subsubsection{Residual Training Results}
With the MJX jitter issue mitigated, the residual policy was trained using Proximal Policy Optimization (PPO) on Google Colab's A100 runtime. The training configuration used 2{,}048 parallel environments at 50~Hz control rate, with an episode length of 4 seconds (200 control steps). The policy and value networks were standard multi-layer perceptrons; the actor received a deploy-able observation comprising sensor history, action history, current motor command, the IK-recovered base control, and the target pose, while the critic received an additional privileged observation containing tendon states and ground-truth pose error.

The residual action was bounded to $\pm 20$~mm per cable per control step (approximately 28\% of the cable stroke range), allowing the policy to make corrections of meaningful magnitude while constraining it from overriding the base controller entirely. Targets were sampled from a cubic distribution over the 300mm workspace radius ($d = d_{\max} \cdot u^3$ where $u \sim \mathcal{U}(0,1)$), naturally emphasizing easy targets early in training while retaining full workspace coverage. Domain randomization was applied to the joint encoder readings with a standard deviation $5\times$ the nominal sensor noise to encourage robustness for eventual sim-to-real transfer.

\begin{table*}[H]
    \centering
    \caption{Residual RL training progression over 50M environment steps.}
    \label{tab:rrltrainingprogress}
    \begin{tabular}{rrrr}
        Steps (M) & Reward & Pos err (mm) & Ang err (deg) \\
        \hline
         0  & 265 & 238 & 17.1 \\
        10  & 329 & 218 & 18.7 \\
        20  & 410 & 184 & 14.3 \\
        30  & 454 & 174 & 12.3 \\
        40  & 488 & 159 & 11.9 \\
        50  & 564 & 162 & 11.6 \\
    \end{tabular}
\end{table*}

Training for 50 million environment steps over approximately 5.5 hours produced clear monotonic learning, but did not converge to a deployment-ready policy within this budget. Table 3 summarizes the progression. The policy reduced mean position error by 32\% and mean orientation error by 32\% relative to the initial random policy, while more than doubling the mean episode reward. However, the final mean position error of 162~mm and orientation error of 11.6$^\circ$ remain substantially worse than the base controller alone evaluated from rest, which achieves a median position error of 3.7mm and sub-degree orientation error on reachable
targets. Several factors contribute to the gap between the trained residual and the base controller's standalone performance. First, the residual policy is evaluated under a more demanding distribution than the offline base controller evaluation: episodes begin from a sequence of states encountered during the ramp-in of the base control command rather than from a clean rest equilibrium, and the encoder observations include the $5\times$ amplified noise used during training. Second, continuous-action PPO on an 8-dimensional control problem with stiff cable dynamics has well-documented sample complexity in the hundreds of millions to billions of environment interactions; 50 million steps represents an early point on the learning curve. Third, the reward landscape combines coarse exponential shaping with discrete success bonuses, and the balance between position and orientation terms was not extensively tuned.

The reward and error trajectories indicate that the architecture and training pipeline are functional, learning is occurring monotonically without instability, but additional compute, reward tuning, and curriculum refinement would be required to reach the level of accuracy demonstrated by the base controller in isolation. Given that the residual reinforcement learning component is not the primary contribution of this work, and the mechanical and base-controller results stand on their own merit, we report these findings as a proof-of-concept that the residual training infrastructure operates correctly, and leave full convergence to future work.

\begin{figure}[!t]
\centering
\subfloat[]{
	\includegraphics[width=0.45\columnwidth]{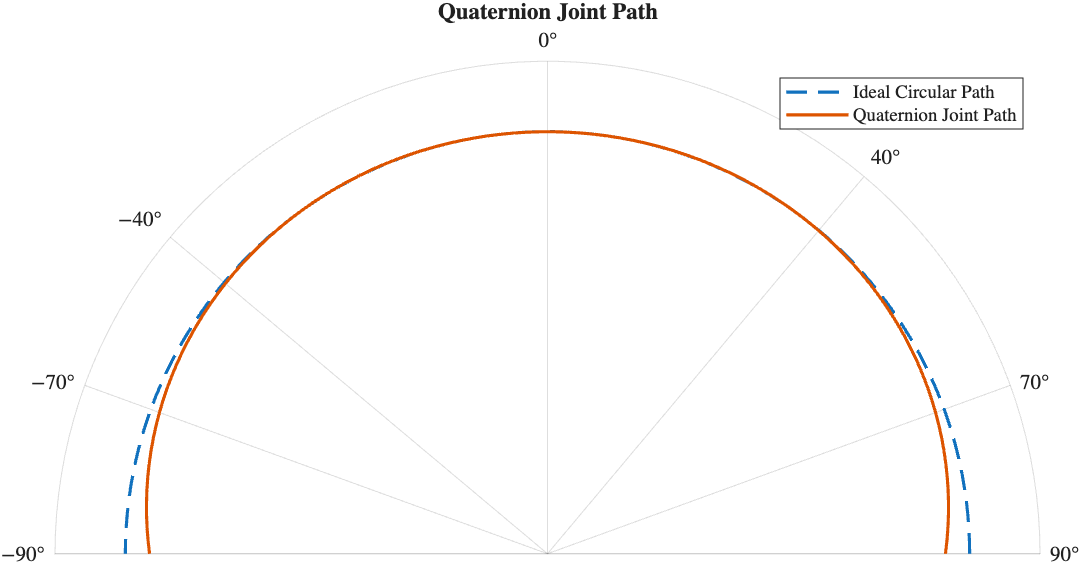}}
\subfloat[]{
	\includegraphics[width=0.45\columnwidth]{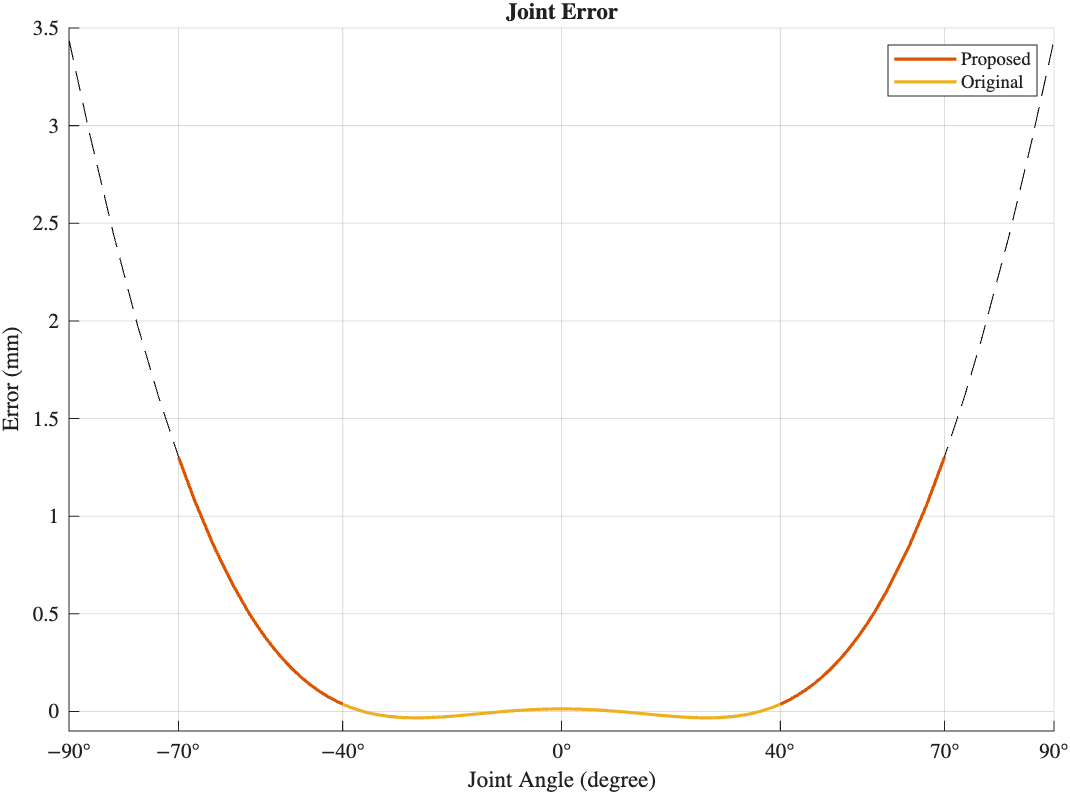}}
\caption{(a) Ideal circular path (dashed blue) overlayed on actual elliptical path (solid red); (b) Deviation (mm) from ideal circular path in function of joint angle; proposed system (red) overlayed on \cite{} (yellow).}
\label{fig:path}
\end{figure}

\subsection{Elliptical Deviation}

The joint model for the joint assumes an ideal spherical motion. This section analyzes the deviation of actual quaternion joint's kinematic path from this ideal, as this error is a critical factor in the manipulator's precision. To quantify this deviation, The polar form of the joint's path is used to compare with the ideal circular path. The resulting error is the positional difference between these two paths. \\

Fig. \ref{fig:path} provides a visual comparison. The quaternion joint path follows the ideal circular path closely at small bending angles. However, as the angle increases, the quaternion joints path begins to curve inward, diverging from the ideal circle. This graph clearly quantifies the observation and reveal two distinct operational regions:
\begin{itemize}
    \item \textbf{Low-Error Region ($[-40\degree, 40\degree]$):} Within the original angle range, the kinematic error is minimal.
    \item \textbf{High-Error Region ($>|40\degree|$):} Beyond $\pm40\degree$, the error grows rapidly.
\end{itemize}
This analysis is directly relevant to our proposal of extending the joint range to $[-70\degree, 70\degree]$. As shown in Fig. \ref{fig:path}, operating at $\pm70\degree$ introduces a predictable, systematic error. \\

Our primary control strategy, which utilizes Residual Reinforcement Learning (RRL), is designed to handle such nonlinearities.

\section{Related Work}\label{sec:related}

Quaternion joints were first introduced by Kim et al \cite{kim2018quaternion} and quickly adopted in several designs \cite{pang2022design, pang2024stiffness}. Their inverse kinematics have been the subject of significant study, including \cite{huang2023sensing,han2026enhanced}.\par Cable-driven mechanisms  have been widely applied in robotics as an alternative to rigid linkages and gear systems \cite{kozak2006static}. By using tensioned cables such as wires, ropes, or tendons to transmit force and motion, they offer several key advantages: lightweight construction, flexible routing, and remote actuation that reduces inertia \cite{lau2013generalized}. In the context of confined servicing \cite{wei2026embodied}, Kim et al \cite{kim2018quaternion} have shown that quaternion joint designs eable an N-segment manipulator can be controlled with just 2N motors, reducing overall complexity while maintaining large bending angles. They employed  a fast, geometric inverse kinematics (IK) solver based on the Forward and Backward Reaching Inverse Kinematics (FABRIK \cite{aristidou2011fabrik}) algorithm for trajectory planning. Incrementally, Huang et al \cite{huang2023sensing} demonstrated the integration of direct joint-state feedback using mechanical encoders. To achieve high-precision motion, they implemented a classical control strategy combining a kinematic feedforward term with a Proportional-Integral-Derivative feedback loop. The experiments successfully validated the system's ability to perform high-precision trajectory tracking.
\par For robotics tasks involving intricate dynamics \cite{zang2016applications}, traditional model-based control becomes exceedingly difficult \cite{kommey2025compact}. Reinforcement Learning (RL) \cite{kober2013reinforcement} has emerged as a powerful, data-driven paradigm for developing controllers in such scenarios. RL agents can learn complex, non-linear control policies directly from interaction with their environment, without requiring a perfect analytical model of the system dynamics \cite{tang2025deep}. Applying RL to physical robots presents significant challenges. Pure RL often suffers from poor sample efficiency \cite{yarats2021improving}, requiring millions of interactions to learn a task, which is impractical on real hardware \cite{hester2013texplore}. To bridge the gap between the stability of classical control and the adaptability of pure RL, the hybrid framework of Residual Reinforcement Learning was introduced by to robotics \cite{johannink2019residual}. This state-of-the-art approach addresses the core challenges of sample efficiency and safety in robotic learning \cite{zhang2024residual}. The key insight is to structure the final control command as a superposition of a conventional, model-based controller \cite{doya2002multiple} and a learned RL policy. The conventional controller provides a stable baseline behavior, handling the coarse motion and ensuring the robot behaves reasonably at all times. The RL agent is then tasked with learning only the "residual" or corrective action \cite{lambert2019low}. This residual policy learns to compensate for all the complex, unmodeled dynamics that the simple base controller cannot handle \cite{m2023model}.  By providing a strong prior from the base controller, this method dramatically improves sample efficiency, as the agent only needs to learn the subtle, corrective part of the policy.

\section{Conclusions}\label{sec:conclusions}

 We have shhown  that a 4-segment, 8-joint manipulator can achieve a broader workspace than extant configurations, at lower hardware cost, and that Residual Reinforcement Learning outperforms extant state-of-the-art methods -specifically, the FABRIK algorithm- on the control of such manipulator. An ongoing challenge is that the complexity of the kinematic model of quaternion joints challenges \textit{a priori} decisions on manipulator configurations and imposes higher computational demands on the control system and its non-linearities amplify all discrepancies between design and physical artifact arising from fabrication imprecision.   Comparative analysis demonstrated that the Learned IK approach outperforms FABRIK by approximately three orders of magnitude in both positional and orientation accuracy, validating the necessity and effectiveness of learning-based methods for compensating real-world mechanical non-linearities.

\bibliography{sample}

\end{document}